\documentclass{article}

\usepackage{PRIMEarxiv}

\usepackage[utf8]{inputenc} 
\usepackage[T1]{fontenc}    
\usepackage{hyperref}       
\usepackage{url}            
\usepackage{booktabs}       
\usepackage{amsfonts}       
\usepackage{nicefrac}       
\usepackage{microtype}      
\usepackage{lipsum}
\usepackage{graphicx}
\graphicspath{{media/}}     
\usepackage{amsmath} 
  
\title{Lightweight Backbone Networks Only Require
Adaptive Lightweight Self-Attention Mechanisms
}

\author{
  Fengyun Li, Chao Zheng$^*$, Yangyang Fang, Jialiang Lan, Jianhua Liang, Luhao Zhang, Fa Si \\
  School of Computer Science and Engineering \\
  Northeastern University \\
  Shenyang, China\\
  \texttt{lifengyun@mail.neu.edu.cn}, \\
  \texttt{\{2372103, 2471996, 2372009, 2301904, 2472086, 2401895\}@stu.neu.edu.cn} \\
}

\begin{document}
\maketitle

\begin{abstract}
Currently, lightweight hybrid backbone networks have partially alleviated the issue of computational saturation, but the imbalance in computational efficiencys between convolutional neural networks (CNNs) and attention mechanisms is becoming increasingly apparent. Specifically, although linear attention mechanisms and their variants have made progress in lightweight design, they still fail to meet the demands of hybrid models for long-sequence modeling. On the other hand, existing lightweight SoftMax attention computations typically reduce the feature map to a fixed size to decrease the number of sequences, thereby compressing the computational scale. However, the process of determining the feature map reduction ratio is cumbersome, and computational saturation issues still persist. To address this issue, this paper proposes a lightweight SoftMax attention mechanism with adaptive feature map sizes, named Fast Window Attention (FWA), which generates a small number of key sequences (Key and Value) through window aggregation for attention computation. Additionally, it explains the rationality of using ReLU to simulate SoftMax operations in lightweight global attention mechanisms. Finally, the paper designs a global-local feature fusion mechanism and combines it with GhostNet to propose a lightweight hybrid backbone network, LOLViT. Through visual tasks such as classification (ImageNet 1K), detection (COCO 2017), and segmentation (BDD100K), along with extensive ablation studies, it is demonstrated that LOLViT outperforms CNN models of the same level in both inference speed and model accuracy. Notably, the inference speed of LOLViT-X is 5× that of MobileViT-X.
\end{abstract}


\section{Introduction}
SoftMax Attention (SA) significantly enhances performance in lightweight hybrid backbone networks. However, high training costs and slow inference speeds make hybrid backbone networks like MobileViT~\cite{01mehta2021mobilevit} challenging to replace lightweight CNNs in practical applications. To improve the inference speed of hybrid backbone networks, current mainstream solutions can be divided into two categories: (1) Reducing computational complexity through structural innovations: These methods no longer perform global multi-head self-attention (MHSA) computations on the entire feature map. Instead, they adopt specialized approaches such as local attention, window attention, hierarchical structures, or dynamic token selection, performing attention computations only on selected key data. These models excel in fusing global and local information while significantly improving inference speed. For instance, PiT~\cite{02heo2021rethinking}, Mobile-Former~\cite{03chen2022mobile}, EfficientFormer~\cite{04li2022efficientformer}, CvT~\cite{05wu2021cvt}, TokenLearner~\cite{06ryoo2021tokenlearner}, EdgeViTs~\cite{07pan2022edgevits}, HiRi-ViT~\cite{08yao2024hiri}, LeViT~\cite{09graham2021levit},  LaViT~\cite{10zhang2024you}, etc. (2) Lightweight MHSA mechanisms: Since the $\mathcal{O}(N^2)$ time complexity of the original MHSA mechanism is a computational bottleneck, many studies focus on reducing MHSA's computational complexity. For instance, models like MobileViTv2~\cite{11mehta2022separable}, CAS-ViT~\cite{12zhang2024cas}, and EfficientViT~\cite{13cai2022efficientvit} replace traditional MHSA with linear attention mechanisms or approximation methods to achieve lightweight effects. Other works, such as LeViT, XFormer~\cite{14zhao2022lightweight}, and FasterViT~\cite{15hatamizadeh2023fastervit}, reduce computation by adjusting the number of attention heads, lowering feature dimensions, or using hierarchical or multi-scale attention. For structures specifically designed for image restoration tasks, such as P2FEViT~\cite{16wang2023p} and HSA~\cite{17sun2024restoring}, their generalizability to other tasks remains to be validated.

\begin{figure}[!htbp]
\centering
\includegraphics[width=1\linewidth]{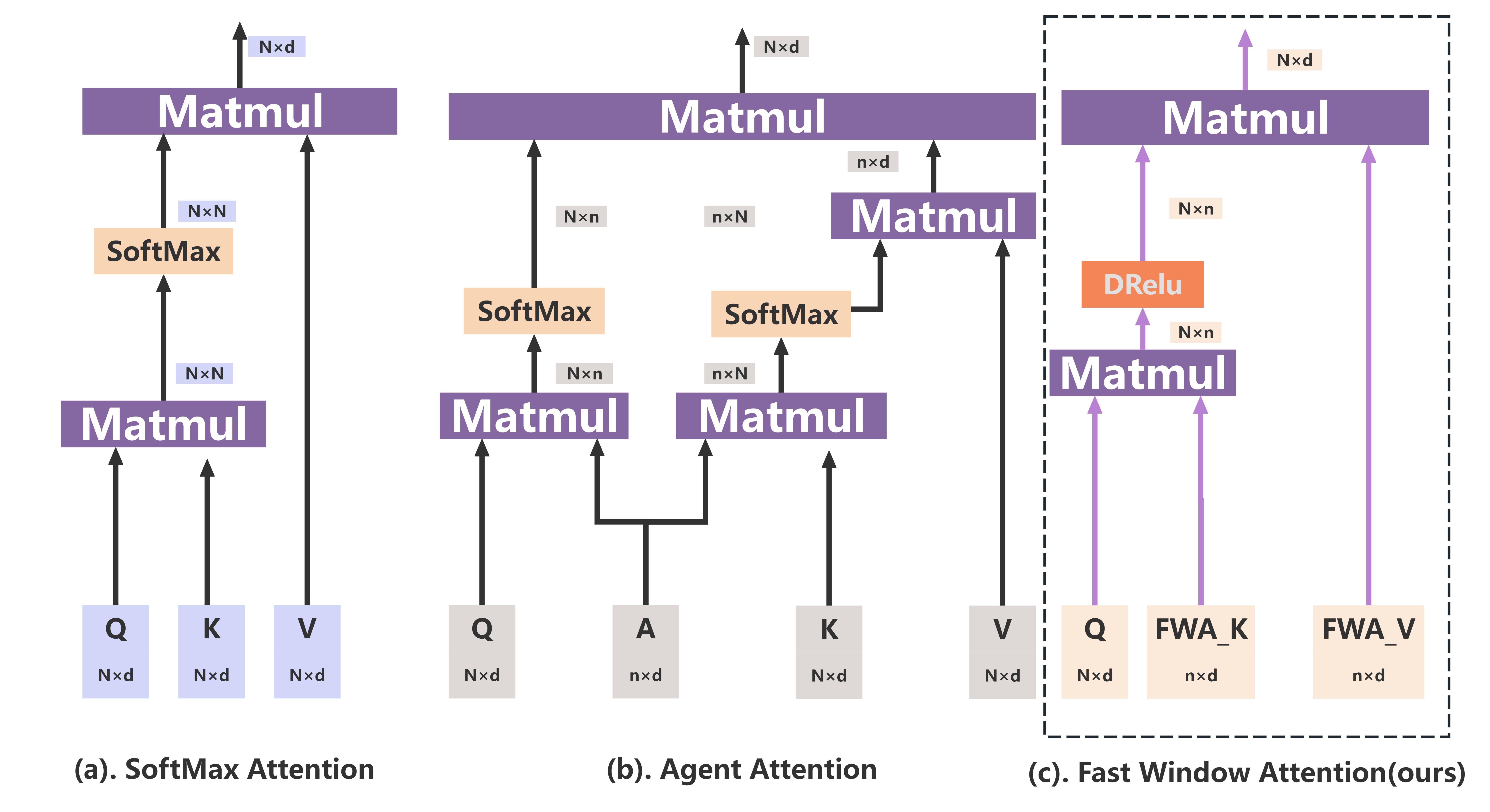}
\caption{\textbf{Difference between Softmax Attention, Agent Attention and Fast Window Attention.} (a) The common Multi-Head Self-Attention mechanism in Transformer, (b) Agent Attention generating key sequences using pooling, (c) The Fast Window Attention mechanism proposed in this paper.}
\label{fig:1}
\end{figure}

Directly applying SA to process raw sequences leads to severe computational saturation issues. Using linear attention mechanisms and their variants, or lightweight SA methods, can significantly reduce the computational costs of training and inference. However, existing studies still have some limitations. For example, the separable attention mechanism in MobileViTv2 and the linear attention mechanism proposed by SLAB~\cite{19guo2024slab} effectively reduce computational complexity but weaken the hybrid model's capability for global modeling of long sequences. Additionally, SA variants like Agent Attention~\cite{18han2024agent} alleviate the computational saturation of multi-head self-attention by introducing Agent key sequences based on QKV, thereby altering the scale of matrix computations. However, Agent sequences are coarse-grained, and their generation method is not CPU-friendly. Moreover, the fixed-length key sequences require repeated hyperparameter tuning for visual tasks involving images of different sizes, increasing the complexity of model training.

To address the computational complexity and performance bottleneck issues faced by existing attention mechanisms, this paper proposes a new lightweight attention mechanism—Fast Window Attention (FWA, Figure~\ref{fig:1}). FWA combines the efficiency of linear attention mechanisms with the accuracy advantages of SA, and introduces the Full-Scene Adaptive Window Aggregation (FAWA) method, which dynamically adjusts the number of key sequences based on the size of the input feature map. Compared to traditional attention mechanisms, FWA maintains high efficiency while generating fine-grained key sequences through simple computations, avoiding complex format conversion operations and significantly improving computational efficiency. Furthermore, the adaptability of FWA allows it to flexibly handle different input sizes, demonstrating strong performance across various vision tasks.

To fully unleash the potential of FWA and enhance the inference speed of hybrid models, this paper further optimizes FWA. First, in the computation of attention weights, the ReLu activation function is used instead of SoftMax to reduce the risk of overfitting in lightweight networks and improve computational efficiency. Second, caching techniques are employed to avoid redundant computations, significantly reducing the overhead caused by repetitive calculations. Finally, a convolutional neural network (CNN) is introduced after the attention computation to capture local information within patches, achieving effective fusion of global and local information and further enhancing model performance.

This paper proposes a lightweight hybrid backbone network -- LOLViT, based on GhostNet, integrating FWA and the designed optimization methods. LOLViT is the first lightweight hybrid backbone network to combine QK caching, simulated SoftMax, and dynamic matching of data processing capabilities between CNN and attention mechanisms. By effectively integrating convolution and attention mechanisms, LOLViT enhances computational efficiency while maintaining excellent feature representation capabilities.

The main contributions of this paper are as follows:
\begin{itemize}
    \item We propose the FWA lightweight attention mechanism, which automatically adapts to feature map sizes and adjusts the number of generated fine-grained key sequences.
    \item We built upon FWA, design DReLu, key sequence caching, and global-local feature fusion methods, significantly improving the inference speed and performance of hybrid backbone networks.
    \item We propose LOLViT, a lightweight hybrid backbone network that dynamically balances the data processing capabilities of convolutional neural networks and attention mechanisms.
    \item We design and conduct experiments on datasets including ImageNet 1K, Mini ImageNet, COCO2017, PASCAL VOC (2012+2007), and BDD100K to validate LOLViT's capabilities, with extensive ablation studies confirming the rationality of the proposed designs.
\end{itemize}

\section{Related Works}
\textbf{Depthwise Separable Convolution.} MobileNetv4~\cite{21qin2024mobilenetv4} and GhostNetv3~\cite{22liu2024ghostnetv3} based on depthwise separable convolutions are the latest representatives of pure CNN architectures. The former employs knowledge distillation to enable lightweight models to learn from larger models, while the latter introduces a parallel network structure and uses reparameterization techniques to convert the parallel network into a single network layer, enhancing inference speed. Additionally, RepViT~\cite{23wang2024repvit} integrates efficient designs from lightweight ViTs, gradually applying them to MobileNetv3~\cite{24howard2019searching}, achieving high accuracy and low latency. ConvMixer~\cite{25trockman2022patches} processes patches used in Transformers~\cite{20vaswani2017attention} as input with standard convolutions, yielding results superior to ViT~\cite{27dosovitskiy2020image}, MLP-Mixer~\cite{28tolstikhin2021mlp}, their variants, and traditional visual models like ResNet~\cite{26he2016deep}. Lightweight convolutional neural networks, constrained by parameter count and network depth, struggle to capture long-range dependencies.

\textbf{Hybrid Network.} Balancing performance and inference speed in hybrid models is challenging: when ViT modules are complex, the backbone network leans toward ViT characteristics, as seen in MobileViT, which underperforms on small datasets compared to equivalent pure CNNs, is sensitive to data augmentation, has high training difficulty, and exhibits relatively slow inference speeds. Conversely, models with linear attention mechanisms, such as MobileViTv2 and MobileViTv3~\cite{33wadekar2022mobilevitv3}, exhibit CNN characteristics but have reduced global information learning capabilities. Research on EfficientFormer indicates that MHSA increases training costs but is not necessarily the most time-consuming component in hybrid backbone networks. Holistic optimization and design of hybrid backbone networks can effectively improve their inference speed. Studies like TinyViT~\cite{34wu2022tinyvit} attempt to apply knowledge distillation techniques in hybrid backbone networks to learn from complex models, enhancing model performance. Recent research on LaViT notes computational saturation in attention mechanisms and demonstrates that caching attention scores can improve model capabilities without frequently recomputing attention weights.

\section{Preliminaries}
\subsection{Lightweight Hybrid Backbone}
To enhance the capabilities of lightweight convolutional neural network models, incorporating attention mechanisms into CNN has become a common practice. To ensure the lightweight nature of hybrid models, most studies introduce attention mechanisms at positions with smaller feature map sizes, such as MobileViT, which employs Transformers in the last three layers of the MobileNet feature pyramid to improve model performance.
\subsection{SoftMax Attention}
The traditional SoftMax Attention (Figure~\ref{fig:1}a) calculates the correlation matrix using the raw Query and Key, then applies SoftMax to obtain the attention weights, and finally performs matrix multiplication with the Value to complete the attention calculation, see Eq. (\ref{eq:1}), with a computational complexity of $\mathcal{O}(N^2D)$. For lightweight backbone networks, directly serializing the feature map into a MHSA for attention computation can lead to computational saturation, resulting in the training and inference costs of the hybrid backbone network being much higher than those of convolutional neural networks.
\begin{equation}
    OutPut = \rm{SoftMax}\left( {\frac{{\mathbf{Q} \times {\mathbf{K}^\mathsf{T}}}}{{\sqrt {\rm{\textit{d}}} }}} \right) \times \mathbf{V}
\label{eq:1}
\end{equation}

Where $\mathbf{Q}, \mathbf{K}, \mathbf{V}$ are the matrices corresponding to Query, Key, and Value sequences  respectively, and $d$ represents the embedding dimension of $\mathbf{Q}, \mathbf{K}, \mathbf{V}$ as the scaling factor for the correlation matrix calculation.

\section{LOLViT}
\begin{figure*}[!t]
\centering
\includegraphics[width=1\linewidth]{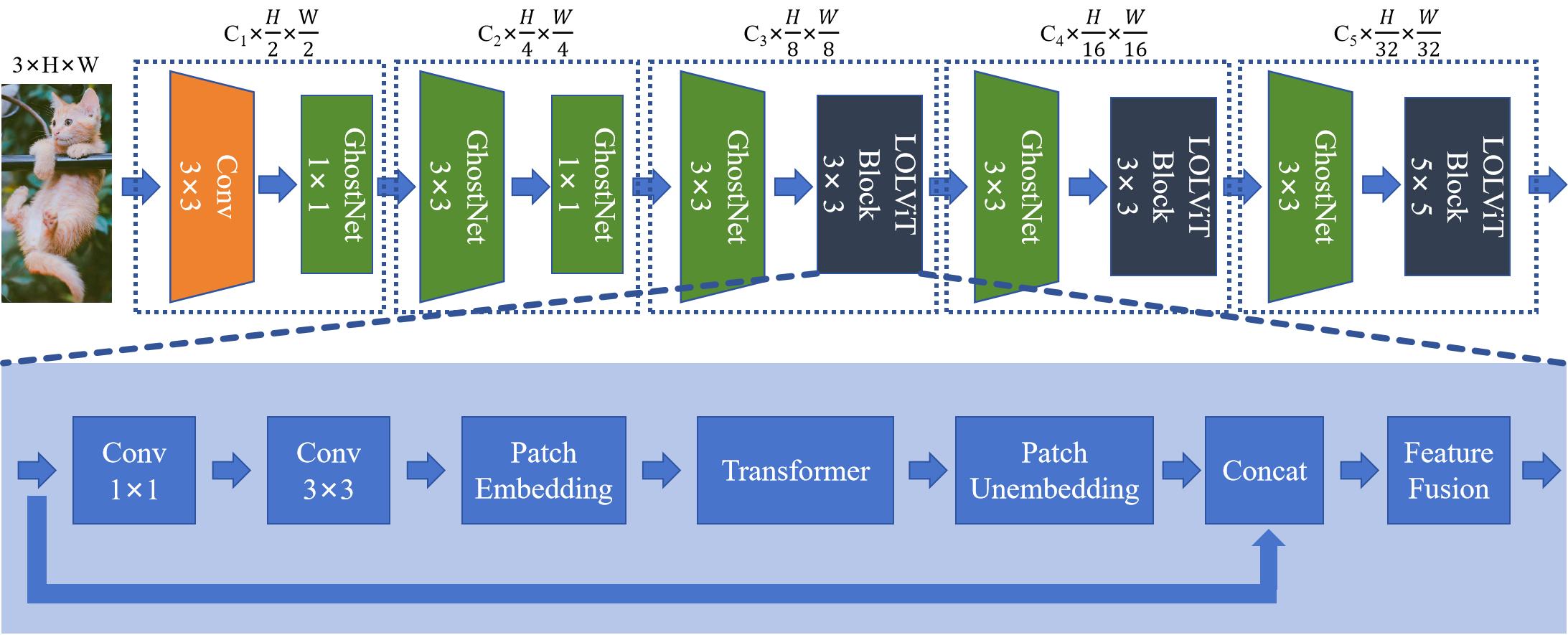}
\caption{\textbf{The architecture of LOLViT proposed in this paper.} LOLViT consists of five network modules, with the last three being ViT modules.}
\label{fig:2}
\end{figure*}

To address the imbalance in computational efficiency between lightweight convolutional neural networks and self-attention mechanisms, this paper uses GhostNet~\cite{35han2020ghostnet} as the baseline and integrates the proposed attention mechanism to introduce LOLViT, an adaptive lightweight hybrid backbone network (Table~\ref{tab:1}). LOLViT employs a 32× image upsampling strategy with a five-layer network structure, where the first two layers are pure CNNs, and the last three layers are hybrid modules combining CNN and ViT. In the hybrid modules, GhostNet extracts features, followed by a convolutional layer for upscaling, and then the feature maps are serialized for attention computation using a Transformer (Figure~\ref{fig:2}).

\begin{table}[!htbp]
\caption{\textbf{Different sizes of LOLViT models}}
\centering
\linespread{1.2}\selectfont
\begin{tabular}{ccccc}
\toprule
Model    & Params(M) & FLOPs(G) & Top1(\%) \\ 
\midrule
LOLViT-X & 1.9       & 9.2      & 72.53    \\
LOLViT-S & 1.1       & 5.9      & 67.43     \\ 
\bottomrule
\end{tabular}
\label{tab:1}
\end{table}

\begin{figure*}[hb]
\centering
\includegraphics[width=1\linewidth]{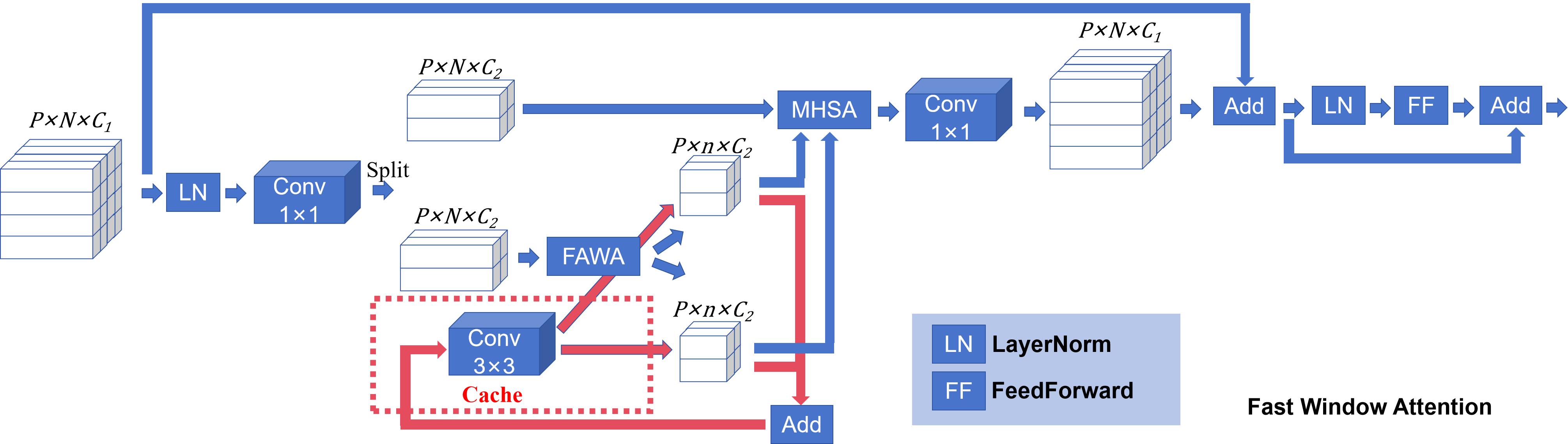}
\caption{\textbf{Fast Window Attention processing pipeline}, where FAWA is the Full-scene Adaptive Window Aggregation method proposed in this paper, MHSA is the multi-head self-attention mechanism, and the SoftMax is replaced with the DReLu method in this study.}
\label{fig:3}
\end{figure*}

Unlike most studies (Figure~\ref{fig:3}), the Transformer structure in this paper initially generates only two sequences: Query and Token. After generating the key sequences, a linear transformation is performed through convolution to obtain two key sequences, $\mathbf{FWA\_K}$ and $\mathbf{FWA\_V}$. These are used alongside Query for attention computation, with attention scores calculated using the DRelu method.

To further improve computational efficiency, this paper designs a key sequence caching mechanism. In scenarios requiring repeated Transformer execution, $\mathbf{FWA\_K}$ and $\mathbf{FWA\_V}$ can be directly retrieved from the cache and undergo linear transformation to complete fast attention computation.

\subsection{Fast Window Attention}
FWA consists of three core components: (1) replacing SoftMax with DReLu to improve the speed of attention weight computation; (2) designing an adaptive key sequence method for image size, which significantly reduces the scale of matrix calculations, reduces training costs, and improves training and inference efficiency; (3) using key sequence caching, where the cached key sequence is linearly transformed and reused when stacking ViT modules, as shown in Eq. (\ref{eq:2}).  

\begin{equation}
    OutPut = \rm{DReLu}\left( {\frac{{\mathbf{Q} \times \mathbf{FWA\_K}^\mathsf{T}}}{{\sqrt d }}} \right) \times \mathbf{FWA\_V}
\label{eq:2}
\end{equation}

Where DReLu is the attention weight calculation method proposed in this paper; $\mathbf{FWA\_K}$ and $\mathbf{FWA\_V}$ are the generated key sequences, which are sourced from the cache when stacking ViT modules.

\begin{figure}[!htbp]
\centering
\includegraphics[width=1\linewidth]{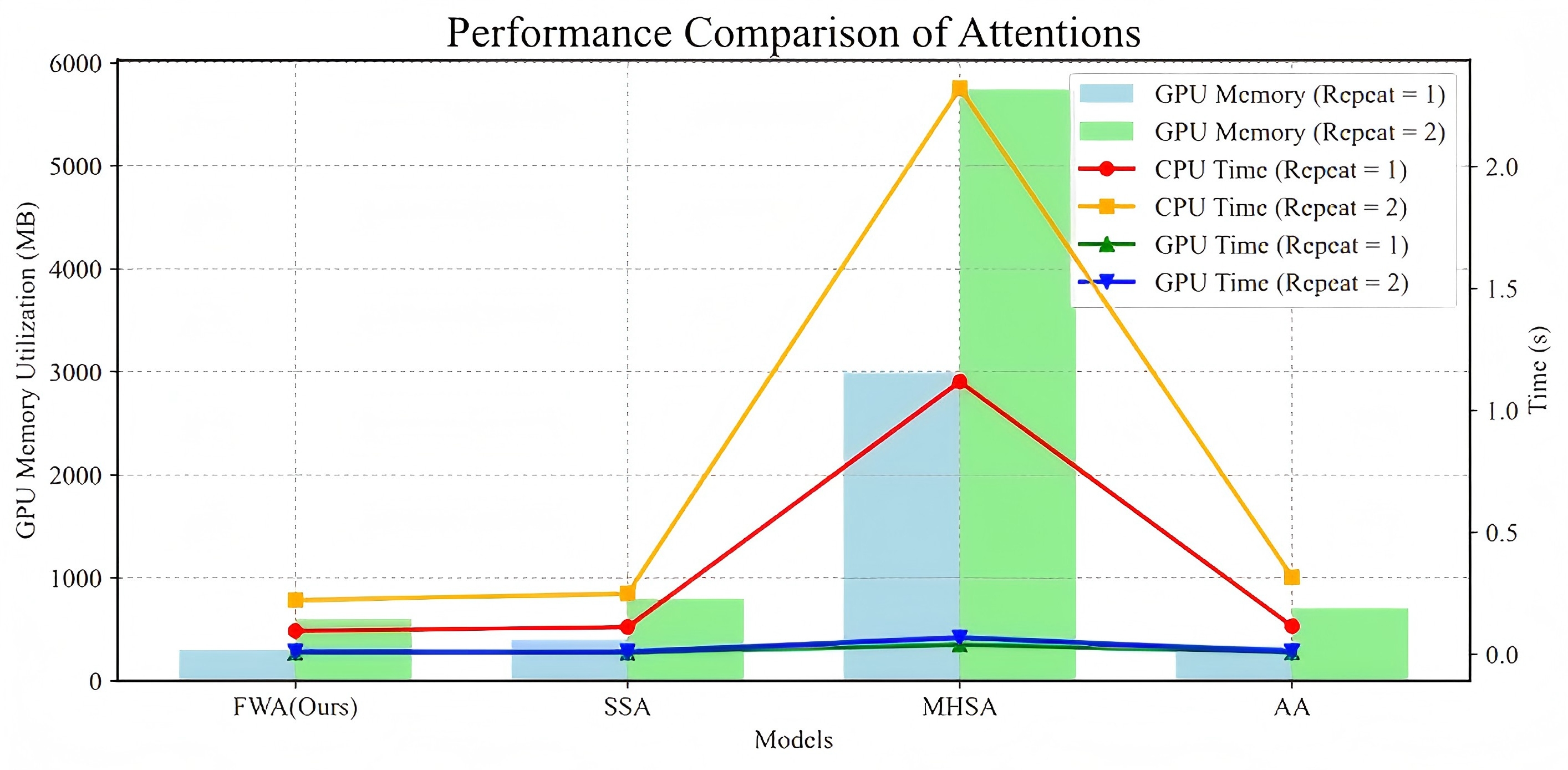}
\caption{\textbf{Comparison of computational efficiency in processing attention mechanisms.} The tensor [128, 96, 32, 32] is processed, and the CPU and GPU processing times, along with memory overhead, are recorded. "Repeat" indicates the number of repetitions of the Transformer structure. FWA: the attention mechanism proposed in this paper, SSA: the separable attention mechanism proposed in MobileViTv2, AA: Agent Attention, MHSA: Multi-Head Self-Attention.}
\label{fig:4}
\end{figure}

\subsection{Dynamic Sequence Length and ReLu}
Mitchell Wortsman et al.~\cite{36wortsman2023replacing} summarized the replacement of SoftMax with ReLu in attention mechanisms, using simple linear computations instead of exponential calculations. Inspired by this, this paper proposes a complete method for correlation matrix calculation using ReLu, called Dynamic Sequence Length and ReLu (DReLu). This paper also provides an explanation of the rationale behind this method in the context of lightweight models: Based on the analysis of the characteristics of visual attention mechanisms, the traditional MHSA module normalizes the correlation matrix of feature maps using the SoftMax function. The core mechanism relies on the strong nonlinear characteristics of exponential operations. However, this nonlinear amplification effect may have a negative impact in lightweight network scenarios: when the network depth is limited and the number of parameters is insufficient, depthwise separable convolution layers struggle to effectively capture the high dynamic range feature differences processed by SoftMax, a phenomenon that becomes particularly pronounced when the number of feature map channels is low.
\begin{figure*}[!htbp]
    \centering
    \begin{minipage}{0.49\linewidth}
		\centering
		\includegraphics[width=0.9\linewidth]{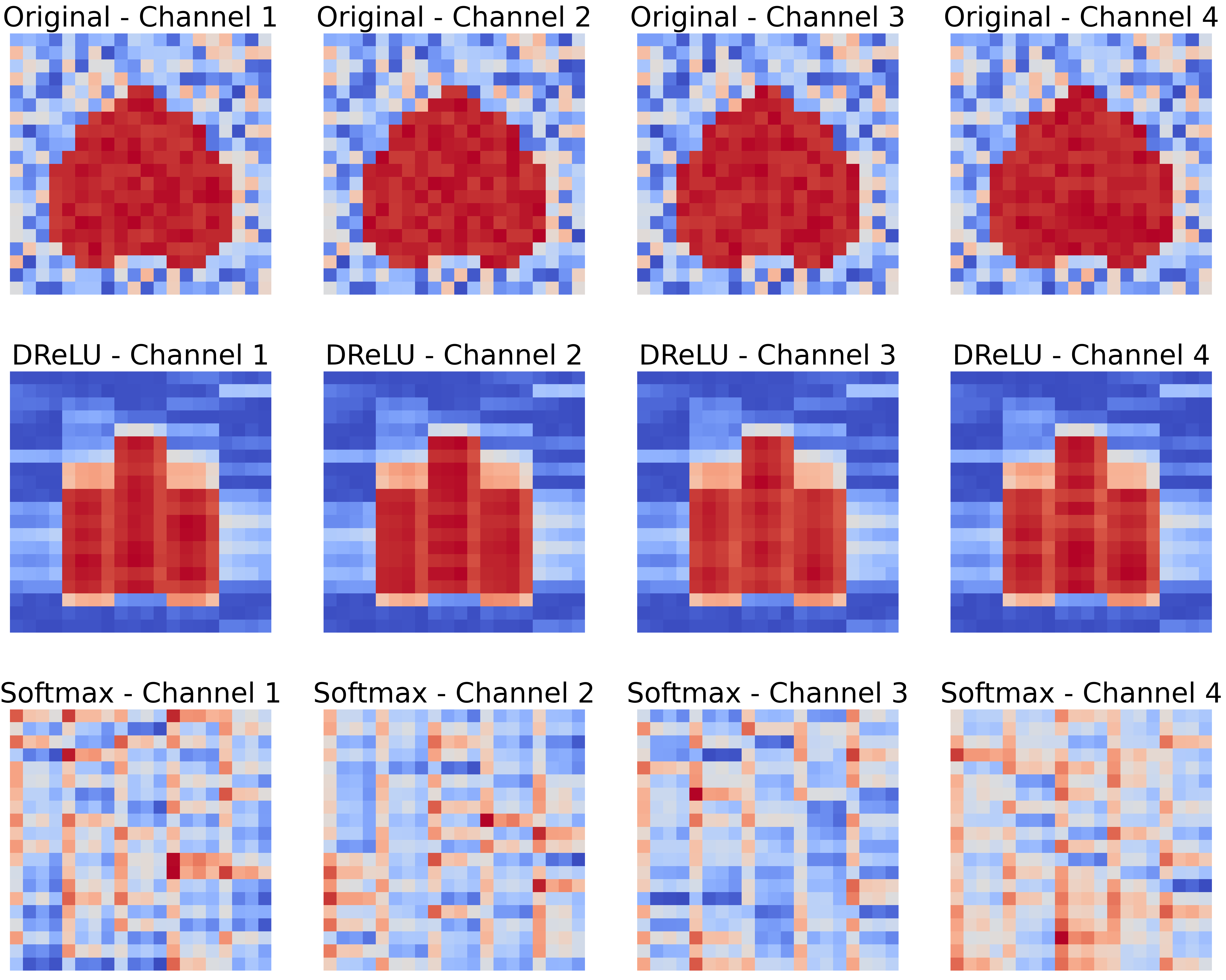}
		\label{fig:6-1}
	\end{minipage}
	\begin{minipage}{0.49\linewidth}
		\centering
		\includegraphics[width=0.9\linewidth]{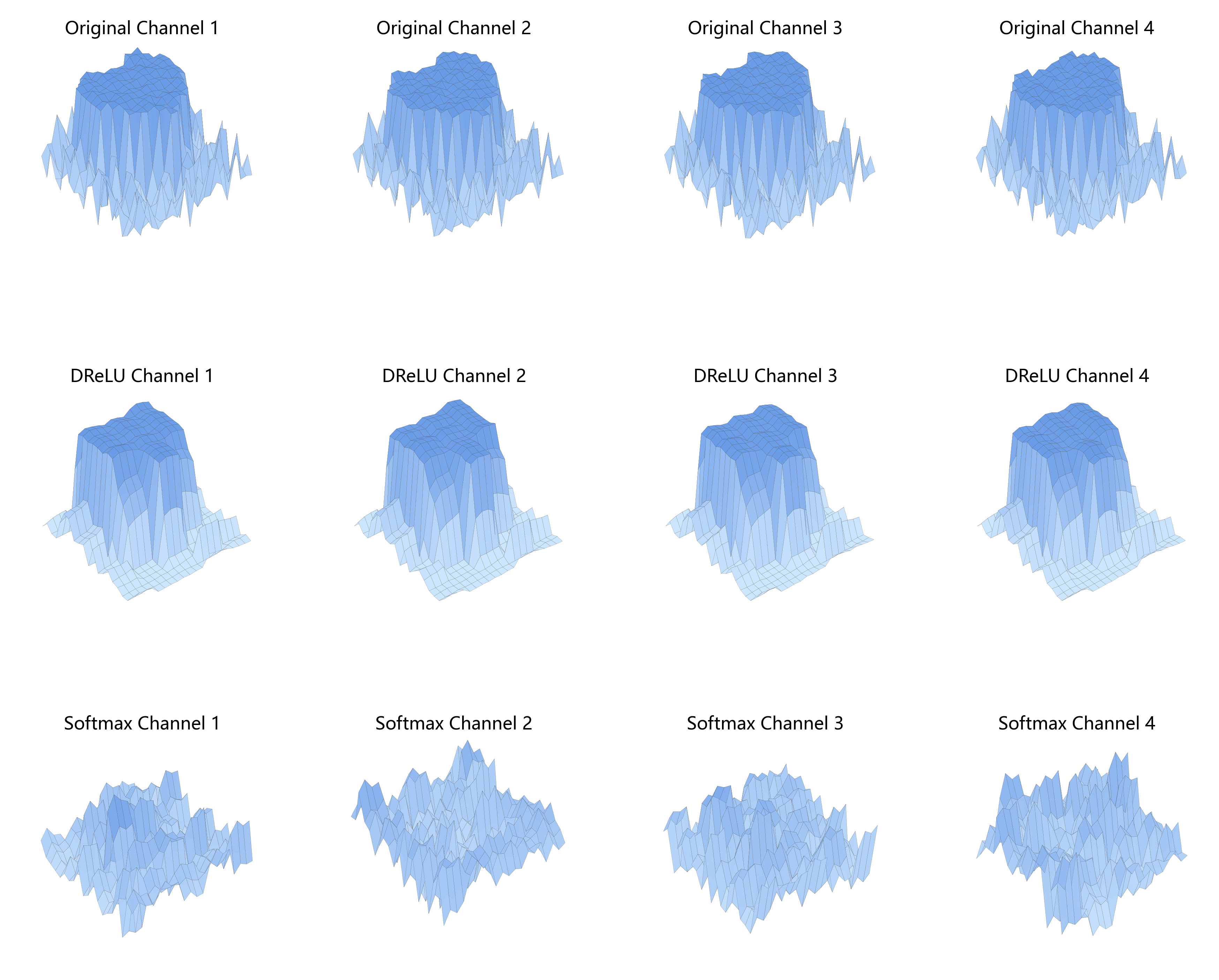}
		\label{fig:6-2}
	\end{minipage}
\caption{\textbf{Feature map visualization using different activation functions.} This paper sets up a 4×20×20 feature map and computes attention weights using DReLu and SoftMax, respectively. The results of the DRelu method clearly distinguish features from the background. Channels 1~4 correspond to these four feature maps.}
\label{fig:5}
\end{figure*}

As shown in the comparative experiment in Figure~\ref{fig:5}, in the four-channel simulation scenario of a 20×20 heart-shaped feature map, the DReLu activation function demonstrates a unique ability to preserve features. This improved activation function inherits the standard ReLU's characteristic of suppressing negative values, while discarding the global normalization mechanism of traditional SoftMax and instead adopting a local feature-adaptive dynamic threshold mechanism. This design allows the feature map to maintain its original distribution characteristics in the spatial dimension, avoiding the issue of feature distribution shift caused by global normalization, and thus forming a better fit with the limited representational capacity of lightweight networks.

From an information theory perspective, the entropy reduction effect introduced by the SoftMax function when calculating attention weights exacerbates information loss in lightweight models. As shown in the visualization results on the right side of Figure~\ref{fig:5}, DReLu preserves the local correlation structure of the feature map, producing more coherent activation patterns in classification tasks, while the feature map processed by SoftMax exhibits overly fragmented activation responses. This contrasts with the limitations of the shallow feature extraction capabilities in lightweight networks.

Experimental validation, as shown in Table~\ref{tab:2}, demonstrates that replacing SoftMax with DReLu in lightweight networks reduces inference time by approximately 1\% and improves accuracy by 0.5\%. The computational process is shown in the Eq. (\ref{eq:3}):
\begin{equation}
    {\left( {\left( {DP + \varepsilon } \right) \cdot L} \right)^{ - 1}} \cdot {\rm{ReLu}}(x)
\label{eq:3}
\end{equation}

where $DP$ is the introduced learnable parameter, $\varepsilon$ is a small inherent bias to prevent the denominator from being zero, and $N$ is the sequence length.
\begin{figure}[!htbp]
    \centering
	\includegraphics[width=1\linewidth]{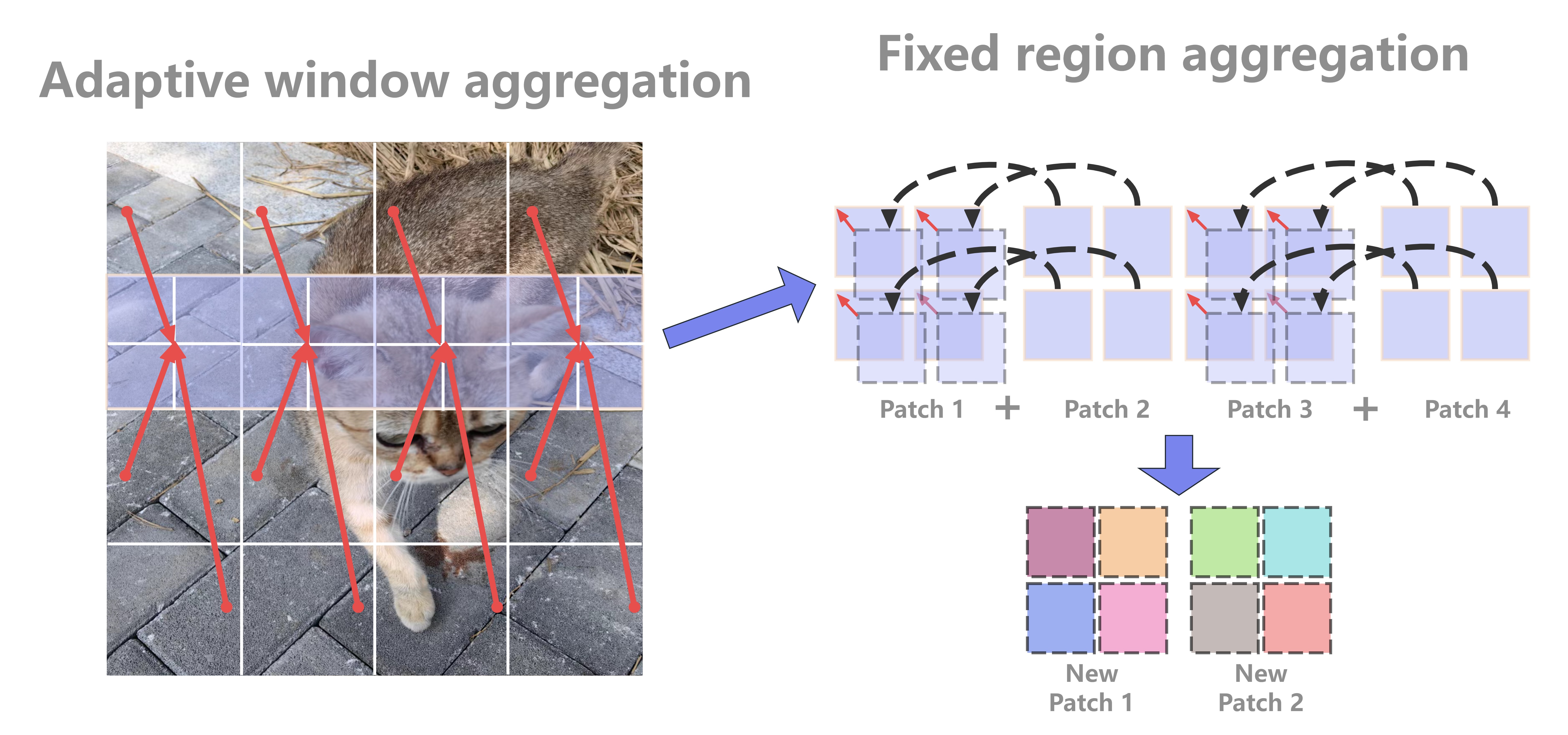}
\caption{\textbf{The key sequence generation method proposed in this paper.} Unlike traditional pooling methods, FAWA does not need to preset the size of the pooled feature map, but adaptively determines the width of the feature map (bounded by the length of the feature map as well as the width of the Patch block). By setting a fixed width of region aggregation (in the figure, this parameter is set to 2, i.e., every two Patches form a fixed region), the sequences are compressed, and the key sequences are finally generated (subject to normalization).}
\label{fig:6}
\end{figure}

\subsection{Full-scene Adaptive Window Aggregation}
The design of FAWA is inspired by Agent Attention (AA). AA reduces the matrix computation scale by using the pooling method to obtain key sequences. During the pooling process, sequences are restored into feature maps, and after pooling, sequences are generated. This process is complex and the generated sequences are coarse-grained. In this paper, we quickly obtain key sequences through some simple calculations. By passing the input image width $I_w$ in the forward pass, and based on the set sequence patch size $P_w$×$P_l$ (where $P_w$ represents the width and $P_l$ represents the length, typically $P_w$=$P_l$), we determine the aggregation window size as $P_l$×$I_w$. The original sequence is then split into $F$ blocks using chunking to form the aggregation windows, and finally, all windows are combined using stacking to generate the key sequence, as shown in Figure~\ref{fig:5}. The computational process is described in Eq. (\ref{eq:4}) (the generated key sequence needs to be normalized before use):
\begin{equation}
    {\mathord{\buildrel{\lower3pt\hbox{$\scriptscriptstyle\frown$}} 
\over w} _i} = \left[ \begin{array}{c}
\sum\limits_{j = i * {P_l}}^{\left( {i + 1} \right) * {{{P}}_l} - 1} {\sum\limits_{t = 0}^{\left\lfloor {\frac{N}{F}} \right\rfloor } {{d_{t * F + j,1}}} } \\
 \vdots \\
\sum\limits_{j = i * {P_l}}^{\left( {i + 1} \right) * {P_l} - 1} {\sum\limits_{t = 0}^{\left\lfloor {\frac{N}{F}} \right\rfloor } {{d_{t * F + j,{{D}}}}} } 
\end{array} \right]
\label{eq:4}
\end{equation}

where  ${\mathord{\buildrel{\lower3pt\hbox{$\scriptscriptstyle\frown$}} 
\over w} _i} \in \mathbb{R}^{1 \times D}$  represents any subsequence in the key sequence($\mathbb{R}^{n \times D}$), $i \in \mathbb{R}^n$, $n$ is the length of the key sequence, $D$ is the head dimension, $F$ is the ratio of the length of the feature map at this network layer to the patch length, $N$ is the original sequence length.

\subsection{Key Sequence Caching}
In the training and inference of large language models, attention weights are typically cached to accelerate the inference process and extend the context window, which has been widely applied in LLM. In ViT, this caching technique also plays a significant role. In the Transformer module, the $\mathbf{Q}$, $\mathbf{K}$, and $\mathbf{V}$ sequences are linear transformations based on the previous attention calculation results. Therefore, by caching the attention weights of $\mathbf{QK}$ and applying linear transformations to the cached weights in the next iteration, the scale of matrix computation can be significantly reduced. To achieve minimized matrix computation scale, this paper retains only the full information of $\mathbf{Q}$ and uses FAWA to perform key sequence extraction for $\mathbf{K}$ and $\mathbf{V}$. However, directly caching the attention weights of $\mathbf{QK}$ would result in a loss of a significant amount of detailed information. To address this, this paper proposes further reducing the computational overhead of FAWA by caching key sequences, see Eq. (\ref{eq:5}). This method effectively reduces computational costs while preserving important features.
\begin{equation}
    \mathbf{FWA\_K,FWA\_V} = \left\{ \begin{array}{l}
Split\left( {Conv\left( {FAWA\left( {Token} \right)} \right)} \right),L = 1\\
Split\left( {Conv\left( {Cache} \right)} \right),L \ge 2
\end{array} \right.
\label{eq:5}
\end{equation}

Where $FAWA(Token)$ is the key sequence generated by aggregating the input sequence through window partitioning. It then passes through a 1×1 convolutional layer and is split into two key sequences, $\mathbf{FWA\_K}$ and $\mathbf{FWA\_V}$, using the Split operation.

\section{Experiments}
\subsection{Image Classification}
This subection verifies the efficiency of the FWA mechanism and the feature extraction capability of the LOLViT hybrid backbone network.

In the experiment shown in Table ~\ref{tab:2}, the number of key sequences used in the AA attention mechanism corresponds to the key sequence length of FWA for 384×384 image size. The experimental results indicate that the AA attention mechanism lacks the ability to adapt to image sizes. For low-resolution images, the key sequence length of AA causes information redundancy, leading to model overfitting. In contrast, FWA successfully avoids this issue, reducing inference time and improving model accuracy.
\vspace{-0.2cm}
\begin{table}[!htbp]
\caption{\textbf{Comparison of the capabilities of different attention mechanisms on Mini ImageNet, CPU-based inference.}}
\centering
\linespread{1.2}\selectfont
\begin{tabular}{cccc}
\toprule
Attention Mechanism & Image Size & Top1(\%) & Infer Time(ms) \\ \midrule
FWA(ours)           & 224*224    & \textbf{82.4 }    & \textbf{12.1}           \\
SSA                 & 224*224    & 81.8     & 12.3           \\
AA                  & 224*224    & 82.2     & 12.6           \\
MHSA                & 224*224    & 82.4     & 13.4           \\
FWA(ours)           & 384*384    & \textbf{84.0}     & 25.2           \\
SSA                 & 384*384    & 83.2     & 25.2           \\
AA                  & 384*384    & \textbf{84.0}     & 25.6           \\
MHSA                & 384*384    & 83.6     & 57.0           \\ \bottomrule
\end{tabular}
\label{tab:2}
\end{table}

\begin{table}[!htbp]
\caption{\textbf{ImageNet 1K classification results, CPU-based inference.}}
\centering
\linespread{1.2}\selectfont
\begin{tabular}{cccccc}
\toprule
Model             & Params(M) & FLOPs(G) & Size    & Top1(\%) & Infer Time(ms) \\
\midrule
LOLViT-X(ours)    & 1.9       & 9.2                                                        & 384*384 & \textbf{74.89}                                               & 25.4           \\
LOLViT-S(ours)    & 1.1       & 5.9                                                        & 384*384 & 69.60                                               & 20.6           \\
LOLViT-X          & -         & -                                                          & 224*224 & 72.53                                               & 12.1           \\
LOLViT-S          & -         & -                                                          & 224*224 & 67.43                                               & \textbf{9.8 }           \\
MobileViT-XS      & 2.3       & 12.8                                                       & 224*224 & 74.80                                               & 46.3           \\
MobileViT-XXS     & 1.4       & 10.7                                                       & 224*224 & 70.18                                               & 43.4           \\
MobileNetv3-small & 2.5       & -                                                          & 224*224 & 66.65                                               & -              \\
Ghostnetv3×0.5    & 2.7       & 10.9                                                       & 224*224 & 69.40                                               & 15.4          \\
\bottomrule
\end{tabular}
\label{tab:3}
\end{table}

The experimental data in Table ~\ref{tab:3} were obtained from extensive training on the ImageNet 1K dataset. The results show that on large datasets, LOLViT has a shorter inference time compared to CNN backbones of the same level, with comparable model performance. MobileViT, benefiting from the global learning ability of MHSA, achieves a significantly higher Top1 score than the LOLViT and CNN models of the same level. However, its balance between inference time and model performance is poor. The inference time of LOLViT-X is only 26\% that of MobileViT-XS. As shown in Table ~\ref{tab:4}, LOLViT performs excellently on the lightweight dataset Mini ImageNet, achieving performance comparable to MobileViT, with only 19\% of its inference time.

The experimental results demonstrate that within small-scale images, LOLViT's ability to adapt to image sizes can effectively improve model accuracy. FWA enables the hybrid model to reach the inference speed of convolutional neural networks. In classification tasks, the model performance surpasses CNN models of the same level.

\begin{table}[!htbp]
\centering
\caption{\textbf{Mini ImageNet classification results, CPU-based inference.}}
\linespread{1.2}\selectfont
\begin{tabular}{cccc}
\toprule
Model          & Size    & Top1(\%) & Infer Time(ms) \\ \midrule
LOLViT-X(ours) & 224*224 & \textbf{82.4}     & \textbf{12.1}           \\
MobileViT-X    & 224*224 & 82.4     & 63.0           \\
MobileViT-XS   & 224*224 & 80.8     & 46.3           \\
MobileViT-XXS  & 224*224 & 77.4     & 43.4          \\ \bottomrule
\end{tabular}
\label{tab:4}
\end{table}
\vspace{-0.3cm}

\subsection{Object Detection}
This subsection verifies the performance of LOLViT in object detection. The datasets used are primarily COCO 2017 (training set: 118,287 images, test set: 5,000 images) and PASCAL VOC (2007+2012) datasets (training set: 21,680 images, test set: 2,700 images).

The experimental data in Table~\ref{tab:5} were obtained using the YOLO Head for training, as the YOLO series by default uses Fuse, meaning that Conv layers are not normalized during inference. To ensure fairness, this paper's model applies Fuse processing to the CNN part during inference, resulting in a 4-8 frame speedup. The experimental results show that LOLViT’s inference speed is close to the YOLO series, as LOLViT’s CNN part heavily utilizes separable convolutions. However, the model’s accuracy remains difficult to surpass compared to the state-of-the-art YOLO models of the same level.
\begin{table}[!htbp]
\centering
\caption{\textbf{Object detection results on COCO 2017 using YOLO Head(640×640), GPU-based inference.}}
\scalebox{0.95}{
\linespread{1.2}\selectfont
\begin{tabular}{cccc}
\toprule
Model            & FLOPs(G) & mAP$^{50\sim95}$(\%) & FPS \\ \midrule
LOLViT-X(Ours)   & 14.5            & \textbf{41.9}              & 99       \\
LOLViT-S(Ours)   & 7.2             & 37.8              & 113      \\
YOLOv5s ~\cite{31YOLOreleasev8.1.0}          & 7.7             & 37.4              & 110      \\
YOLOv8n ~\cite{31YOLOreleasev8.1.0}          & 8.7             & 37.3              & \textbf{134}      \\
YOLOv9t ~\cite{29wang2024yolov9}  & 7.7             & 38.3              & -        \\
YOLOv10n ~\cite{30wang2024yolov10} & 6.7             & 38.5              & -        \\
YOLOv11n ~\cite{31YOLOreleasev8.1.0} & 6.5             & 39.5              & 118      \\
YOLOv12n ~\cite{32tian2025yolov12} & 6.5             & 40.6              & -       \\ \bottomrule
\end{tabular}
}
\label{tab:5}
\end{table}

\begin{table}[!htbp]
\centering
\vspace{-0.3cm}
\caption{\textbf{Object detection results on COCO 2017 using SSD Head(320×320), CPU-based inference.}}
\linespread{1.2}\selectfont
\begin{tabular}{ccc}
\toprule
Model            & mAP & FPS \\ \midrule
LOLViT-X(ours)   & 23.36        & 20.4     \\
MobileViT-XS     & \textbf{24.80}        & 10.1     \\
MobileViTv2-0.5  & 21.24        & \textbf{23.9}     \\
MobileViTv2-0.75 & 24.57        & 14.5    \\ \bottomrule
\end{tabular}
\label{tab:6}
\end{table}

\begin{table}[!htbp]
\centering
\caption{\textbf{Object detection results on PASCAL VOC(2007+2012,640×640) using YOLO Head, CPU-based inference.}}
\linespread{1.2}\selectfont
\begin{tabular}{cccc}
\toprule
Model              & Params(M) & mAP$^{50}$(\%) & FPS \\ \midrule
LOLViT-X(ours)     & 3.6       & 77.5      & 10.9     \\
LOLViT-S(ours)     & \textbf{1.6 }      & 74.5      & \textbf{24.5}     \\
MobileViT-XS       & 2.2       & 77.9      & 5.4      \\
MobileViT-XS(FWA)  & -         & 77.4      & \textbf{9.6(↑78\%)}   \\
MobileViT-X        & 5.19      & \textbf{79.3}      & 2.4      \\
MobileViT-X(FWA)   & -         & 78.6      & \textbf{5.9(↑156\%)}   \\
YOLOv8n            & 3.2       & 74.8      & 21.4     \\
YOLOv8s            & 11.2      & 78.8      & 13.5     \\
YOLOv6s            & 16.31     & 79.0      & -        \\
YOLOv5s            & 9.13      & 77.4      & 10.5     \\
GhostNetv2 ~\cite{37tang2022ghostnetv2} & 5.06      & 76.9      & 10.4     \\
MoblieNetv3        & 3.08      & 71.9      & 12.7     \\
FasterNet ~\cite{38chen2023run}  & 4.11      & 76.2      & -       \\ \bottomrule
\end{tabular}
\label{tab:7}
\end{table}

To demonstrate the generalizability of this model, training was conducted using the SSD Head, and the experimental results are shown in Table 6. The inference speed of LOLViT lies between MobileViTv2 ×0.5 and 0.75, with a 5\% reduction in accuracy compared to MobileViT, and an approximate 2× speedup in inference (FPS). The results in Table 6 prove that at low-resolution scales, LOLViT and MobileViT of the same size have similar accuracy, but LOLViT offers a significant speed advantage. Table 7 indicates that for the smaller VOC dataset (640×640), MobileViT with global attention has a clear accuracy advantage. In a CPU environment, LOLViT’s inference speed is close to that of CNN models of similar size, with a noticeable accuracy improvement. Compared to hybrid models, when trained with large-scale images, FWA will lose some key information, leading to a decrease in accuracy. However, due to the rapid aggregation mechanism of FWA, it ensures a significant increase in the inference speed of the hybrid model with only a limited reduction in accuracy.

For detection tasks, the LOLViT series using the FWA attention mechanism combines the advantages of ViT and CNN in inference tasks for small-scale image object detection. Its inference speed is close to that of CNN models of the same level, and its accuracy is close to that of MobileViT models of the same level. When trained with large-scale images, LOLViT still maintains an advantage over CNN models of the same level.
\subsection{Instance Segmentation}
This section of the experiment is designed to verify LOLViT's capabilities in instance segmentation tasks. A total of 11,000 images were selected from the BDD100K dataset (9,000 training images and 2,000 test images) for instance segmentation, using a multi-task head~\cite{40wu2022yolop} for training.

As shown in Table~\ref{tab:8}, because A-YOLOM-n ~\cite{39wang2024you} uses YOLOv8n as the backbone network, LOLViT slightly lags behind in terms of FPS. However, LOLViT demonstrates stronger performance in terms of mIoU for lane lines and drivable areas. LOLViT has found an excellent balance between pure CNN and hybrid models, with an FPS of 490, close to CNN, while the mIoU is closer to that of the hybrid model.
\begin{table}[!htbp]
\centering
\caption{\textbf{Instance segmentation results on BDD 100k(1/10, 224×224)  using A-YOLOM Head, GPU-based inference.}}
{
\linespread{1.2}\selectfont
\begin{tabular}{ccccc}
\toprule
Model          & Params(M) & mIoU(lane) & mIoU(drivable) & FPS \\ \midrule
A-YOLOM-n      & 3.63      & 0.621      & 0.894          & \textbf{550}          \\
LOLViT-S(ours) & 1.6       & \textbf{0.623}      & 0.895          & 490          \\
MobileViT-XXS  & 1.44      & 0.622      & \textbf{0.896}          & 270         \\ \bottomrule
\end{tabular}
}
\label{tab:8}
\end{table}
\vspace{-0.45cm}

\subsection{Summary of Experiments}
After experimental analysis of the three major computer vision tasks—classification, object detection, and instance segmentation—it was found that LOLViT maintains excellent performance on small-scale images (320×320 COCO 2017) and small-scale datasets (Mini-ImageNet), especially on platforms with limited computational power (CPU). However, on large-scale datasets (COCO, VOC, ImageNet 1K), LOLViT does not have the accuracy advantage over hybrid models (MHSA), with FPS close to 5× that of MobileViT-X.

\section{Ablation Study}
To validate the rationality of the structure proposed in this paper, detailed ablation experiments were conducted on each component. The datasets used for the ablation experiments are Mini ImageNet and ImageNet 1K. The experiments focus on the basic parameters of the Transformer structure in this paper, the learnable parameters in DReLu, as well as ReLu and its variants, the number of heads in the multi-head attention mechanism in the three Transformer modules of the LOLViT network structure, cache, and finally, the FeatureFusion mechanism proposed in this paper.

\textbf{ReLu and its variants.} As shown in Table~\ref{tab:9}, to explore the impact of ReLu and its variants on the DReLu method, this paper compares ReLu and its variants. The model is trained using the parameter set L(1-1-1), Fold(1-2-4), Patch size(1-1-1), with the best experimental results obtained using ReLu.
\begin{table}[!htbp]
\centering
\caption{\textbf{Comparison of ReLu and its variants on ImageNet 1K.}}
\linespread{1.2}\selectfont
\begin{tabular}{ccccc}
\toprule
Way           & \textbf{ReLu} & LeakyReLu & ReLu6 &SiLU \\ \midrule
Top1(\%)      & \textbf{58.0} & 57.5      & 57.6  & 57.6 \\ \bottomrule
\end{tabular}
\label{tab:9}
\end{table}

\textbf{Learnable parameters (DP).} DReLu adds learnable parameters to obtain dynamic sequence lengths, which are then used to scale the correlation matrix. To achieve the best results, this paper sets up ten experiments with a step size of 0.1 for the learnable parameters. It was found that the optimal performance is achieved when DP=0.5 from Table~\ref{tab:10}.
\begin{table}[!htbp]
\centering
\caption{\textbf{The learnable parameter DP ablation experiment of DReLu.}}
{
\linespread{1.2}\selectfont
\begin{tabular}{ccccccccccc}
\toprule
DP       & 0.1  & 0.2  & 0.3  & 0.4  & 0.5  & 0.6  & 0.7  & 0.8  & 0.9  & 1    \\ \midrule
Top1(\%) & 57.5 & 57.4 & 56.7 & 57.2 &\textbf{ 58.0} & 57.9 & 56.9 & 57.5 & 57.7 & 57.9  \\ \bottomrule
\end{tabular}
}
\label{tab:10}
\end{table}

\textbf{Global-local feature fusion.} As shown in Table~\ref{tab:11}, this paper compares FeatureFusion with the feature fusion method used in MobileViTv1. The simple parallel network structure design of FeatureFusion outperforms MobileViTv1, with a significant reduction in both parameter count and computational complexity, while also providing a slight improvement in model accuracy.
\begin{table}[!htbp]
\centering
\caption{\textbf{FeatureFusion on Mini ImageNet.}}
\linespread{1.2}\selectfont
\begin{tabular}{cccc}
\toprule
Model               & Params(M) & FLOPs(G) & Top1(\%) \\ \midrule
FeatureFusion(ours) & 4.5       & 13.1     & 84.4     \\
MobileViTv1         & 5.2       & 15.3     & 84.0    \\ \bottomrule
\end{tabular}
\label{tab:11}
\end{table}

\textbf{The number of Heads.} In the multi-head attention mechanism, the number of heads also impacts the model's performance. As shown in Table~\ref{tab:12}, this paper sets up multiple experiments to explore the optimal number of heads for LOLViT. The experimental results indicate that setting the Transformer Head in LOLViTBlock to 4 yields the best accuracy. Head(2-4-6) denotes the number of attention heads in each LOLViTBlock layer (first:2, second:4, third:6).
\begin{table}[!htbp]
\centering
\caption{\textbf{Layer-wise Head Configuration Analysis in LOLViT (ImageNet-1K).}}
\linespread{1.2}\selectfont
\begin{tabular}{cccccc}
\toprule
Head     & 2-4-6 & 4-6-8 & 4-4-4 & 2-2-2 & 6-6-6 \\ \midrule
Top1(\%) & 57.6  & 57.5  & 58.0  & 57.0  & 57.9  \\ \bottomrule
\end{tabular}
\label{tab:12}
\end{table}

\textbf{Cache.} As shown in Table~\ref{tab:13}, to further reduce the model's inference time, this paper caches the key sequences and, when repeatedly executing the Transformer modules, applies a convolutional neural network linear transformation to the cached key sequences before performing attention computation. The model is trained using the parameter set L(2-4-3), Fold(1-2-4), Patch size(1-1-1), and the experimental results show that the caching method designed in this paper effectively reduces inference time. 

L(2-4-3) denotes the overlapping counts in the three-layer LOLViTBlock architecture (first:2, second:4, third:3). Each ViT structure performs secondary compression on key sequences according to Fold size to obtain smaller key sequences. Patch size indicates the dimension of individual sequences within each ViT layer.
\begin{table}[!htbp]
\centering
\caption{\textbf{Cache Configuration in LOLViT (Mini ImageNet 384×384), CPU-based inference.}}
\linespread{1.2}\selectfont
\begin{tabular}{ccc}
\toprule
L-Fold-Patch       & Top1(\%) & Infer Time(ms) \\ \midrule
243-124-111(Cache) & 84.4     & 27.2                 \\
243-124-111        & 84.2     & 28.9                  \\
111-124-111        & 84.0     & 25.2                \\ \bottomrule
\end{tabular}
\label{tab:13}
\end{table}

This paper progressively combines the methods proposed in this work with the baseline network GhostNet.
\begin{table}[htbp]
\centering
\caption{\textbf{Overall ablation experiment results, CPU-based inference.}}
\linespread{1.2}\selectfont
\begin{tabular}{cccccccc}
\toprule
MHSA & FWA & AA & SoftMax & DReLu & FeatureFusion & Top1(\%) & Infer Time(ms) \\ \midrule
×    & ×   & ×  & ×       & ×     & ×             & 82.4     & 26.5           \\
$\surd$    &     &    & $\surd$       &       & $\surd$             & 83.6     & 57             \\
$\surd$    &     &    &         & $\surd$     & $\surd$             & 83.4     & 54             \\
     &     & $\surd$  & $\surd$       &       & $\surd$             & 84.0     & 25.6           \\
     &     & $\surd$  &         & $\surd$     & $\surd$             & 83.7     & 25.6           \\
     & $\surd$   &    & $\surd$       &       & $\surd$             & 83.6     & 25.4           \\
     & $\surd$   &    &         & $\surd$     & ×             & 83.6     & 25.2           \\
     & $\surd$   &    &         & $\surd$     & $\surd$             & \textbf{84.0}     & \textbf{25.2}      \\ \bottomrule    
\end{tabular}
\label{tab:14}
\end{table}

Overall, as shown in Table~\ref{tab:14}, the attention mechanism, feature fusion mechanism, and DReLu method designed in this paper successfully enhance the feature extraction capability of the baseline network while effectively reducing inference time.
 
\section{Conclusion}
To address the computational efficiency balance between CNNs and attention mechanisms in lightweight hybrid backbone networks, this paper proposes a lightweight global attention mechanism with adaptive feature scales, FWA. FWA incorporates efficient designs such as DReLu, FAWA, and key sequence caching, achieving state-of-the-art (SOTA) computational efficiency for global attention mechanisms. Using GhostNet as the CNN component, this paper constructs a lightweight hybrid backbone network, LOLViT. Through experiments on classification, detection, and segmentation tasks using authoritative datasets, LOLViT demonstrates superior accuracy and inference speed compared to lightweight convolutional neural network models of the same level, making it a viable alternative to lightweight pure convolutional models.


\bibliographystyle{unsrt}  
\bibliography{references}  

\clearpage
\section*{Appendix}
\subsection*{A. FeatureFusion}
\begin{figure}[!hbp]
  \centering
  \includegraphics[width=1\linewidth]{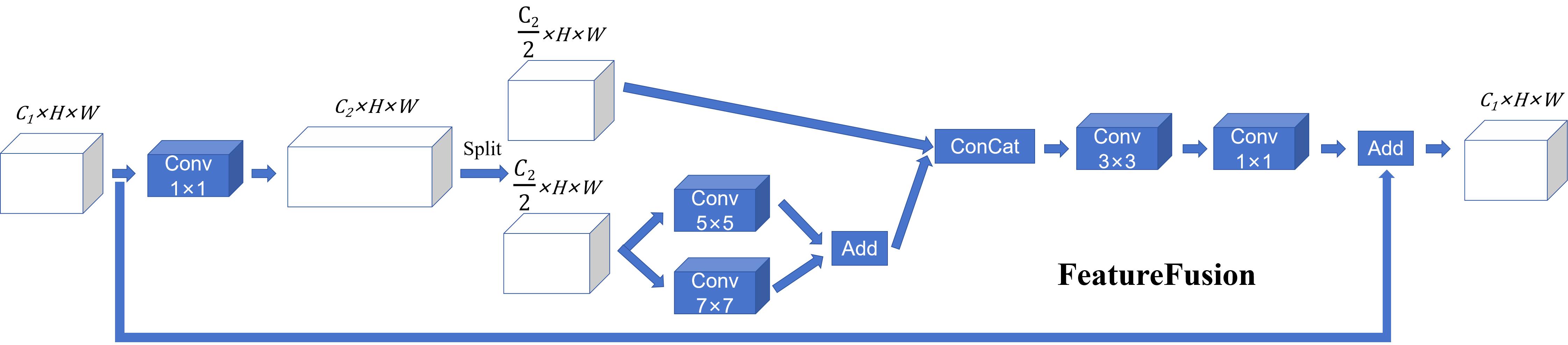}
      \caption{\textbf{FeatureFusion processing process.}}
  \label{fig:7}
\end{figure}
Multi-Head Self-Attention (MHSA) calculates the correlation between each patch and other patches, integrating global information, which allows the model to capture long-range dependencies and contextual features within the image. However, MHSA has limited ability to learn local information within each patch. The FeedForward Network processes each position of all patches independently, without relying on other positions in the sequence. By applying the same nonlinear transformation to the features of each position, it enables the model to capture global semantic information and achieve consistent feature representations across all positions.

Since both of the primary components of the Transformer cannot effectively learn fine-grained local features within each patch, this paper proposes a solution by designing a small parallel network structure on the standard bottleneck architecture. This network layer uses large convolution kernels, such as 5×5 and 7×7, to provide richer feature maps at a very low cost (Figure~\ref{fig:7}).

\subsection*{B. FAWA VS Pool}
\begin{figure}[!hbp]
  \centering
  \includegraphics[width=0.5\linewidth]{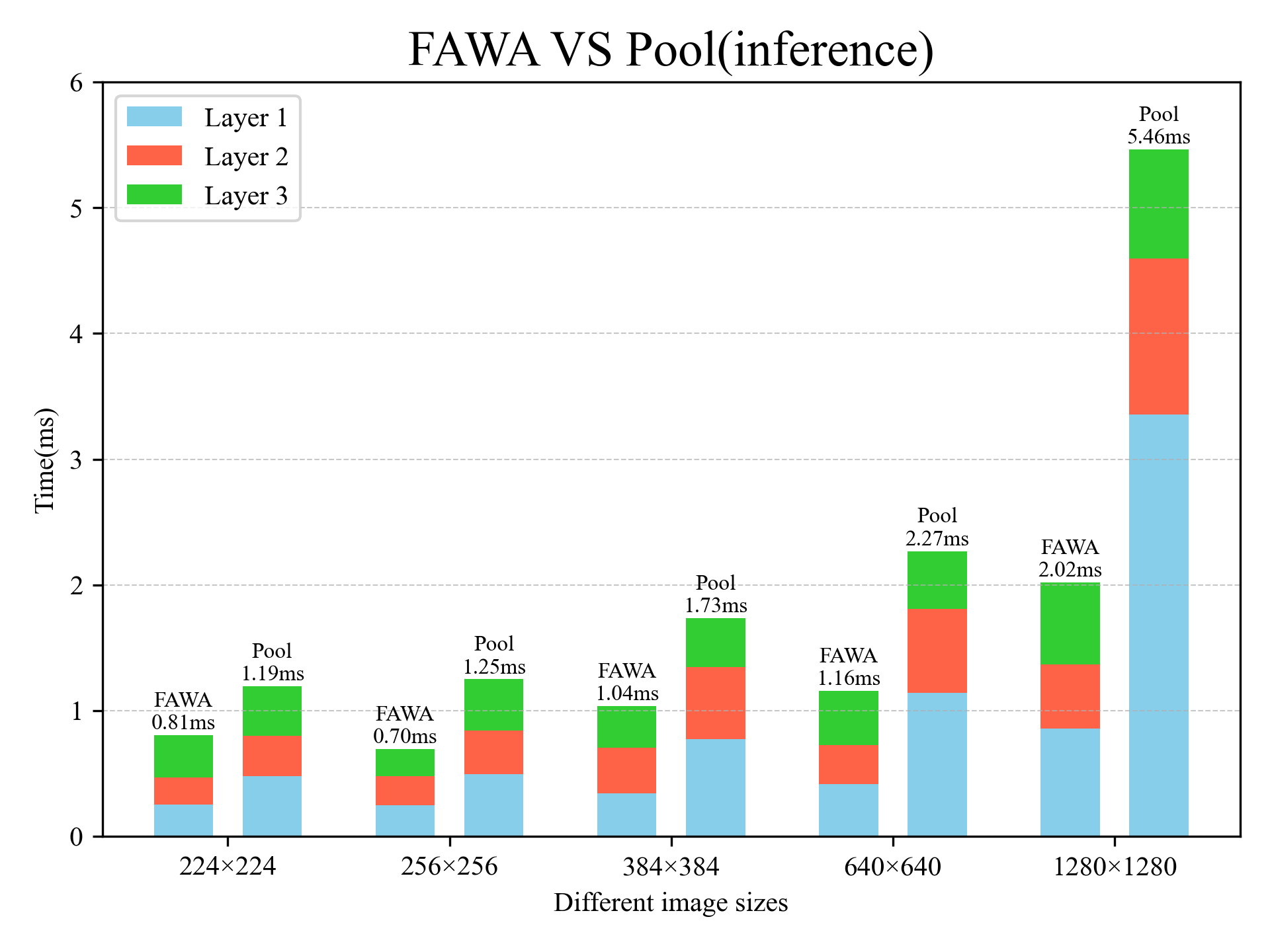}
      \caption{\textbf{Comparison of inference cost between FAWA and pooling method for tasks of different sizes (Intel i5 13600KF CPU).} Layer 1 to Layer 3 correspond to the feature map processing at sizes of 1/8, 1/16, and 1/32 in the backbone network. The total processing time for these three network layers is then measured. FAWA consistently processes faster than the pooling method across tasks of different scales.}
  \label{fig:8}
\end{figure}

The fast aggregation mechanism of FAWA avoids the complex sequence conversion process. The actual efficiency is shown in Figure~\ref{fig:8}. When processing images of different scales, FAWA consistently maintains an advantage.For large-scale images, FAWA's processing speed is approximately 2.7 times faster than the pooling method. The experimental results demonstrate the feasibility of replacing the pooling method with FAWA.

\subsection*{C. DReLu}
To obtain the best model, this paper trains each set of parameters for 100 epochs on Mini ImageNet using a 1*3090 GPU. In the experimental groups with different values of L, Fold, Patch, and Way, the MHSA exceeds the single 3090 GPU memory capacity when using a 384*384 input size, while other cells marked as "-" indicate that no training was conducted. The experimental results (Table~\ref{tab:9}) show that the DReLu method can effectively replace the SoftMax method, significantly reducing inference time. Moreover, the FWA (SoftMax) method results in shorter inference times in most cases.
\begin{table}[!htbp]
\centering
\caption{\textbf{Supplementary experiments of DReLu.}}
\scalebox{0.75}{
\linespread{1.2}\selectfont
\begin{tabular}{cccccc}
\toprule
L-Fold-Patch & Way & Image & Batch & Top1(\%) & Times(ms) \\
\midrule
222-222-421 & FWA(DReLu) & 224*224 & 32 & \textbf{82.7} & \textbf{13.3} \\
222-222-421 & FWA(SoftMax) & 224*224 & 32 & 82.4 & 13.5 \\
222-222-421 & MHSA & 224*224 & 32 & 82.6 & 13.4 \\
222-222-421 & AA & 224*224 & 32 & 82.5 & 14.1 \\
\midrule
222-222-421 & FWA(DReLu) & 384*384 & 32 & 83.6 & 27.4 \\
222-222-421 & FWA(SoftMax) & 384*384 & 32 & 83.6 & \textbf{27.3} \\
222-222-421 & MHSA & 384*384 & 32 & \textbf{83.9} & 27.9 \\
222-222-421 & AA & 384*384 & 32 & 83.7 & 27.5 \\
\midrule
111-111-221 & FWA(DReLu) & 224*224 & 32 & 82.6 & 12.4 \\
111-111-221 & FWA(SoftMax) & 224*224 & 32 & 82.2 & 12.6 \\
111-111-221 & MHSA & 224*224 & 32 & \textbf{82.7} & \textbf{12.3} \\
111-111-221 & AA & 224*224 & 32 & 82.4 & 12.6 \\
\midrule
111-111-221 & FWA(DReLu) & 384*384 & 32 & 83.5 & 25.3 \\
111-111-221 & FWA(SoftMax) & 384*384 & 32 & 83.0 & \textbf{25.2} \\
111-111-221 & MHSA & 384*384 & 32 & 83.7 & 34.1 \\
111-111-221 & AA & 384*384 & 32 & \textbf{83.8} & 25.5 \\
\midrule
111-124-211 & FWA(DReLu) & 224*224 & 32 & 81.8 & \textbf{12.1} \\
111-124-211 & FWA(SoftMax) & 224*224 & 32 & - & - \\
111-124-211 & MHSA & 224*224 & 32 & \textbf{82.3} & 12.4 \\
111-124-211 & AA & 224*224 & 32 & 82.0 & 12.3 \\
\midrule
111-124-211 & FWA(DReLu) & 384*384 & 32 & 83.3 & \textbf{25.2} \\
111-124-211 & FWA(SoftMax) & 384*384 & 32 & - & - \\
111-124-211 & MHSA & 384*384 & 32 & \textbf{83.8} & 31.2 \\
111-124-211 & AA & 384*384 & 32 & 83.7 & 26.4 \\
\midrule
222-222-421 & FWA(DReLu) & 224*224 & 128 & 80.1 & 13.8 \\
222-222-421 & FWA(SoftMax) & 224*224 & 128 & 80.0 & 13.6 \\
222-222-421 & MHSA & 224*224 & 128 & \textbf{80.3} & \textbf{13.5} \\
222-222-421 & AA & 224*224 & 128 & 79.5 & 14.1 \\
\midrule
222-222-421 & FWA(DReLu) & 384*384 & 128 & \textbf{81.5} & \textbf{27} \\
222-222-421 & FWA(SoftMax) & 384*384 & 128 & 81.4 & 27.5 \\
222-222-421 & MHSA & 384*384 & 128 & - & - \\
222-222-421 & AA & 384*384 & 128 & 81.2 & 28.0 \\
\midrule
111-111-221 & FWA(DReLu) & 224*224 & 128 & \textbf{80.2} & 12.3 \\
111-111-221 & FWA(SoftMax) & 224*224 & 128 & 80.2 & 12.3 \\
111-111-221 & MHSA & 224*224 & 128 & 79.8 & \textbf{12.2} \\
111-111-221 & AA & 224*224 & 128 & 80.1 & 12.3 \\
\midrule
111-111-221 & FWA(DReLu) & 384*384 & 128 & \textbf{81.7} & 25.1 \\
111-111-221 & FWA(SoftMax) & 384*384 & 128 & 81.5 & \textbf{25.0} \\
111-111-221 & MHSA & 384*384 & 128 & - & - \\
111-111-221 & AA & 384*384 & 128 & 81.0 & 25.5 \\
\midrule
111-124-211 & FWA(DReLu) & 224*224 & 128 & \textbf{80.3} & \textbf{12.2} \\
111-111-221 & FWA(SoftMax) & 224*224 & 128 & - & - \\
111-124-211 & MHSA & 224*224 & 128 & 79.7 & 12.5 \\
111-124-211 & AA & 224*224 & 128 & 80.1 & 12.4 \\
\midrule
111-124-211 & FWA(DReLu) & 384*384 & 128 & 81.4 & 25.3 \\
111-111-221 & FWA(SoftMax) & 384*384 & 128 & - & - \\
111-124-211 & MHSA & 384*384 & 128 & - & - \\
111-124-211 & AA & 384*384 & 128 & - & - \\
\bottomrule
\end{tabular}}
\label{tab:15}
\end{table}

\subsection*{D. Basic structure of the model}
Tables~\ref{tab:16} and \ref{tab:17} present the detailed parameter tables for the LOLViT X and S models in this paper. The parameter count of LOLViT-S is only 1.1M, while LOLViT-X has 1.9M parameters. From the perspective of parameter count, LOLViT is highly lightweight.


\begin{table*}[!htbp]
\begin{minipage}{0.49\textwidth}
\centering
\caption{\textbf{The architecture of LOLViT-S.}}
\scalebox{0.8}{
\linespread{1.2}\selectfont
\begin{tabular}{ccccc}
\toprule
Module & Input & Output & Kernel size & Parameters \\
\midrule
Conv & 3 & 32 & 3×3 & 464 \\
GhostNetBlock & 32 & 32 & 1×1 & 2,448 \\
GhostNetBlock & 32 & 64 & 3×3 & 6,872 \\
GhostNetBlock & 64 & 64 & 1×1 & 6,696 \\
GhostNetBlock & 64 & 96 & 3×3 & 17,208 \\
LOLViTBlock & 96 & 96 & -- & 163,927 \\
GhostNetBlock & 96 & 128 & 3×3 & 30,288 \\
LOLViTBlock & 128 & 128 & -- & 295,703 \\
GhostNetBlock & 128 & 192 & 3×3 & 60,528 \\
LOLViTBlock & 192 & 192 & -- & 472,535 \\
\midrule
\multicolumn{4}{c}{Number of model parameters} & 1,056,669 \\
\bottomrule
\end{tabular}}
\label{tab:16}
\end{minipage}
\hfill
\begin{minipage}{0.49\textwidth}
\centering
\caption{\textbf{The architecture of LOLViT-X.}}
\scalebox{0.8}{
\linespread{1.2}\selectfont
\begin{tabular}{ccccc}
\toprule
Module & Input & Output & Kernel size & Parameters \\
\midrule
Conv & 3 & 32 & 3×3 & 928 \\
GhostNetBlock & 32 & 32 & 1×1 & 3,440 \\
GhostNetBlock & 32 & 64 & 3×3 & 8,160 \\
GhostNetBlock & 64 & 64 & 1×1 & 10,976 \\
GhostNetBlock & 64 & 96 & 3×3 & 30,288 \\
LOLViTBlock & 96 & 96 & -- & 269,943 \\
GhostNetBlock & 96 & 128 & 3×3 & 60,528 \\
LOLViTBlock & 128 & 128 & -- & 468,439 \\
GhostNetBlock & 128 & 192 & 3×3 & 109,728 \\
LOLViTBlock & 192 & 192 & -- & 900,279 \\
\midrule
\multicolumn{4}{c}{Number of model parameters} & 1,862,709 \\
\bottomrule
\end{tabular}}
\label{tab:17}
\end{minipage}
\end{table*}

\subsection*{E. Image Classification}
The detailed parameter settings of the image classification experiment are shown in Table~\ref{tab:18} and \ref{tab:19}.


\begin{table*}[!htbp]
\begin{minipage}{0.49\textwidth}
\centering
\caption{\textbf{Detailed parameters of the experimental setup(Table~\ref{tab:2} \& Table~\ref{tab:4}).}}
\scalebox{0.8}{
\linespread{1.2}\selectfont
\begin{tabular}{cc}
\toprule
Item & Value \\
\midrule
Training Platforms & 1* NVIDIA GeForce RTX 3090 \\
Inference Platforms & CPU 13th Gen Intel(R) Core(TM) i5-13600KF \\
Operation System & Ubuntu \\
Batch Size & 32 \\
Image Size & 384 \& 224 \\
Datasets & Mini ImageNet \\
epoch & 100 \\
optimizer & AdamW \\
Learning rate & 0.001 \\
momentum & 0.937 \\
AMP & True \\
Augment & True \\
ultralytics & 8.1.45 \\
\bottomrule
\end{tabular}}
\label{tab:18}
\end{minipage}
\hfill
\begin{minipage}{0.49\textwidth}
\centering
\caption{\textbf{Detailed parameters of the experimental setup(Table~\ref{tab:3}).}}
\scalebox{0.8}{
\linespread{1.2}\selectfont
\begin{tabular}{cc}
\toprule
Item & Value \\
\midrule
Training Platforms & 8* NVIDIA GeForce RTX 3090 \\
Inference Platforms & CPU 13th Gen Intel(R) Core(TM) i5-13600KF \\
Operation System & Ubuntu \\
Batch Size & 128*8 \\
Image Size & 384 \& 224 \\
Datasets & ImageNet 1K \\
epoch & 500 \\
optimizer & SGD \\
Learning rate & 0.01 \\
momentum & 0.9 \\
AMP & True \\
Augment & True \\
ultralytics & 8.1.45 \\
\bottomrule
\end{tabular}}
\label{tab:19}
\end{minipage}
\end{table*}

\subsection*{F. Object Detection}
The detailed parameter settings of the object detection experiment are shown in Table~ \ref{tab:20} and \ref{tab:21}.
\begin{table}[!htbp]
\centering
\caption{\textbf{Detailed parameters of the experimental setup(Table \ref{tab:5} \& Table \ref{tab:7}).}}
\linespread{1.2}\selectfont
\begin{tabular}{cc}
\toprule
Item & Value \\
\midrule
Training Platforms & 8* NVIDIA GeForce RTX 3090 \\
Inference Platforms & NVIDIA GeForce RTX 4060 Ti (16G) \& \\
                    & CPU 13th Gen Intel(R) Core(TM) i5-13600KF \\
Operation System & Ubuntu \\
Batch Size & 32*8 \\
Image Size & 640*640 \\
Datasets & COCO 2017 \& PASCAL VOC \\
epoch & 500 \\
optimizer & SGD \\
Learning rate & 0.01 \\
momentum & 0.9 \\
AMP & True \\
Augment & True \\
ultralytics & 8.1.45 \\
\bottomrule
\end{tabular}
\label{tab:20}
\end{table}

\begin{table}[!htbp]
\centering
\caption{\textbf{Detailed parameters of the experimental setup(Table \ref{tab:6}).}}
\linespread{1.2}\selectfont
\begin{tabular}{cc}
\toprule
Item & Value \\
\midrule
Training Platforms & 8* NVIDIA GeForce RTX 3090 \\
Inference Platforms & NVIDIA GeForce RTX 4060 Ti (16G) \& \\
                    & CPU 13th Gen Intel(R) Core(TM) i5-13600KF \\
Operation System & Ubuntu \\
\midrule
Batch Size & 32*8 \\
Image Size & 320*320 \\
Datasets & COCO 2017 \\
epoch & 300 \\
Same training parameters used with MobileViTv2 & \\
\bottomrule
\end{tabular}
\label{tab:21}
\end{table}


\subsection*{G. Instance Segmentation}
The detailed parameter settings of the instance segmentation experiment are shown in Table~\ref{tab:22}.
\begin{table}[!htbp]
\centering
\caption{\textbf{Detailed parameters of the experimental setup(Table~\ref{tab:8}).}}
\linespread{1.2}\selectfont
\begin{tabular}{cc}
\toprule
Item & Value \\
\midrule
Training Platforms & 1* NVIDIA GeForce RTX 3090 \\
Inference Platforms & NVIDIA GeForce RTX 4060 Ti (16G) \\
Operation System & Ubuntu \\
Batch Size & 32 \\
Image Size & 224*224 \\
Datasets & BDD100K(1/10) \\
epoch & 300 \\
Same training parameters used with A-YOLOM & \\
\bottomrule
\end{tabular}
\label{tab:22}
\end{table}

\subsection*{H. Ablation Experiment}
The detailed parameter settings of the ablation experiment are shown in Table~\ref{tab:23}.
\begin{table}[!htbp]
\centering
\caption{\textbf{Ablation experiment setup.}}
\linespread{1.2}\selectfont
\begin{tabular}{cc}
\toprule
Item & Value \\
\midrule
Training Platforms & 1* NVIDIA GeForce RTX 3090 \\
Inference Platforms & CPU 13th Gen Intel(R) Core(TM) i5-13600KF \\
Operation System & Ubuntu \\
Batch Size & 32 \\
Image Size & 224*224 \\
Datasets & ImageNet 1K \& Mini ImageNet \\
epoch & 20 or 100 \\
optimizer & AdamW \\
Learning rate & 0.001 \\
momentum & 0.937 \\
AMP & True \\
Augment & True \\
ultralytics & 8.1.45 \\
\bottomrule
\end{tabular}
\label{tab:23}
\end{table}

\textbf{The basic parameters of the Transformer structure.} L represents the number of times the Transformer module in the LOLViTBlock (shown in Figure 2) is repeated (2-4-3, corresponding to the three layers of LOLViTBlock). Fold refers to the parameter for further folding the key sequences generated by FAWA. Patch size indicates the size of the tokens (1-1-1, meaning each layer of the LOLViTBlock module has a patch size of 1×1). According to the experimental data in Table 22, it can be observed that increasing L has little effect on the model’s performance. Specifically, the configurations of L(1-1-1), Fold(1-2-4), and Patch size(1-1-1) strike a balance
between performance and model size.

\begin{table}[!htbp]
\centering
\caption{\textbf{Basic hyperparameters of the model on ImageNet 1K.}}
\linespread{1.2}\selectfont
\begin{tabular}{cccc}
\toprule
L & Fold & Patch size & Top1(\%) \\
\midrule
2-4-3 & 1-2-4 & 1-1-1 & \textbf{57.6} \\
2-4-3 & 1-1-1 & 2-2-1 & 57.5 \\
2-2-2 & 1-2-4 & 2-2-1 & 56.9 \\
2-2-2 & 1-1-1 & 2-1-1 & 57.2 \\
1-1-1 & 4-2-1 & 2-2-1 & 56.9 \\
1-1-1 & 2-4-8 & 2-2-1 & 56.9 \\
1-1-1 & 1-2-4 & 2-1-1 & 57.0 \\
1-1-1 & 1-2-4 & 1-1-1 & 57.2 \\
\bottomrule
\end{tabular}
\label{tab:24}
\end{table}

\subsection*{I. Others}
\begin{figure}[!hbp]
  \centering
  \includegraphics[width=1\linewidth]{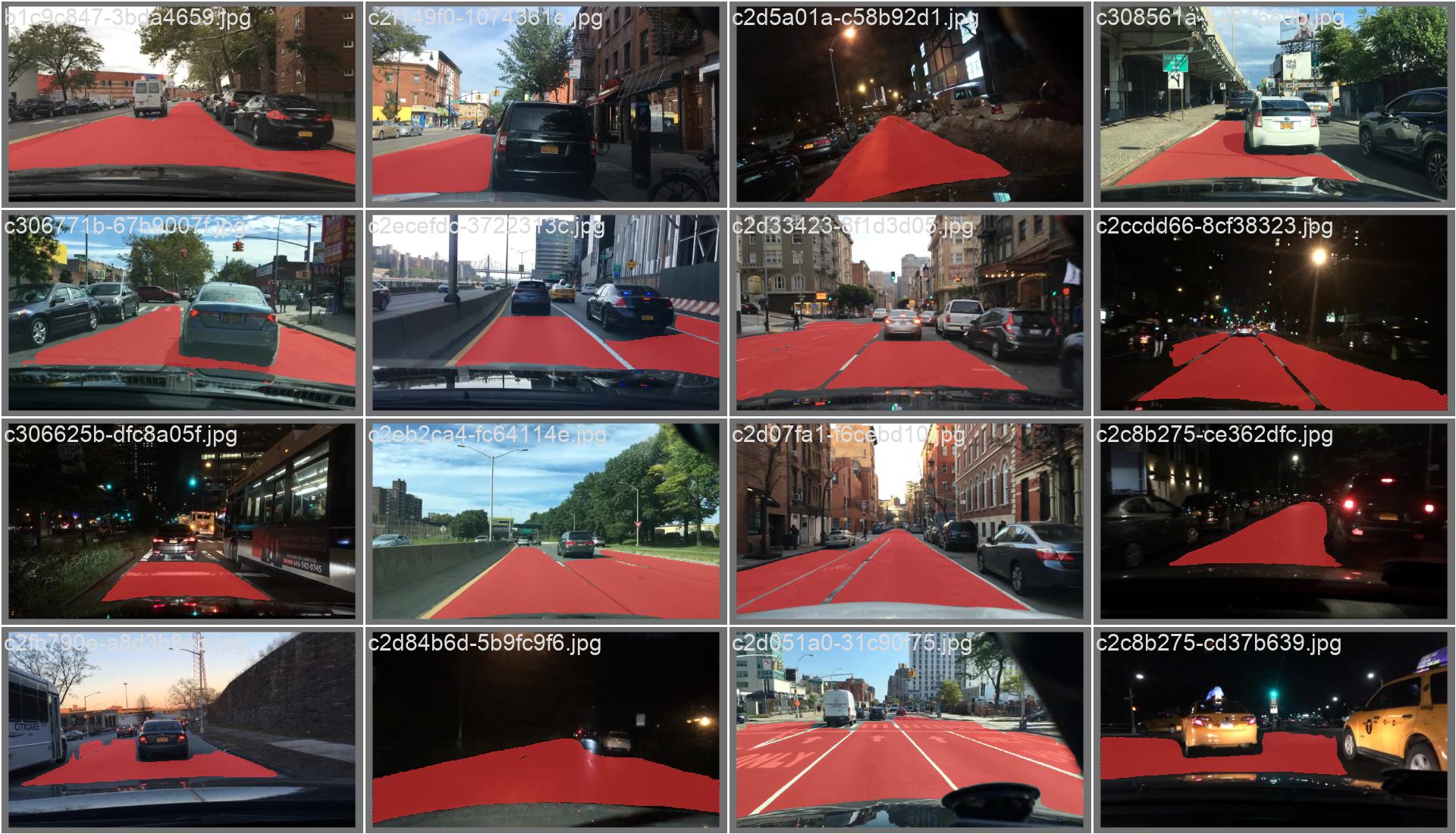}
      \caption{\textbf{Inference results for the division of drivable area (Instance segmentation).}}
  \label{fig:9}
\end{figure}

\begin{figure}
  \centering
  \includegraphics[width=1\linewidth]{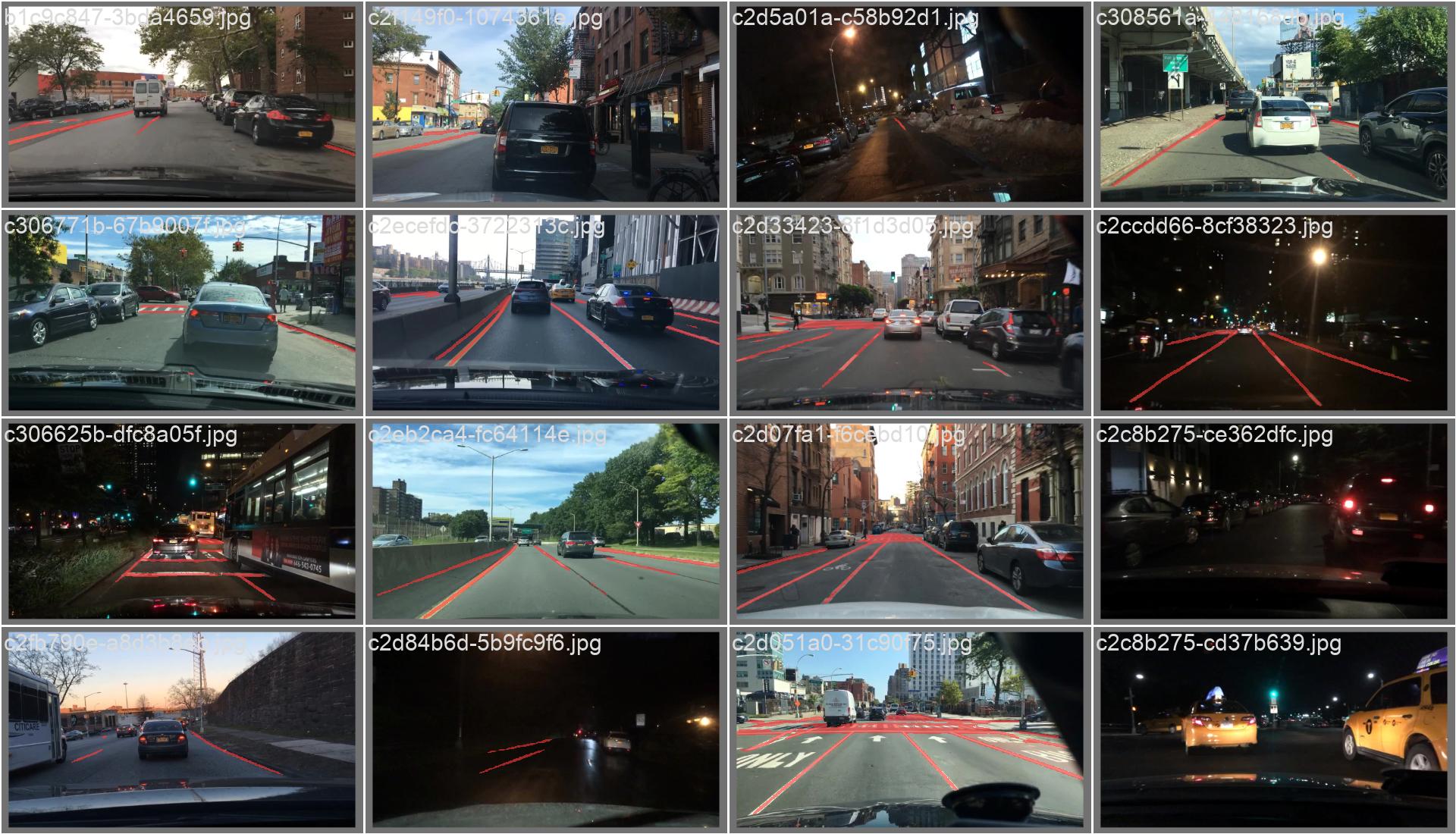}
      \caption{\textbf{Inference results for lane line(Instance segmentation).}}
  \label{fig:10}
\end{figure}

\begin{figure}[!hbp]
  \centering
  \includegraphics[width=1\linewidth]{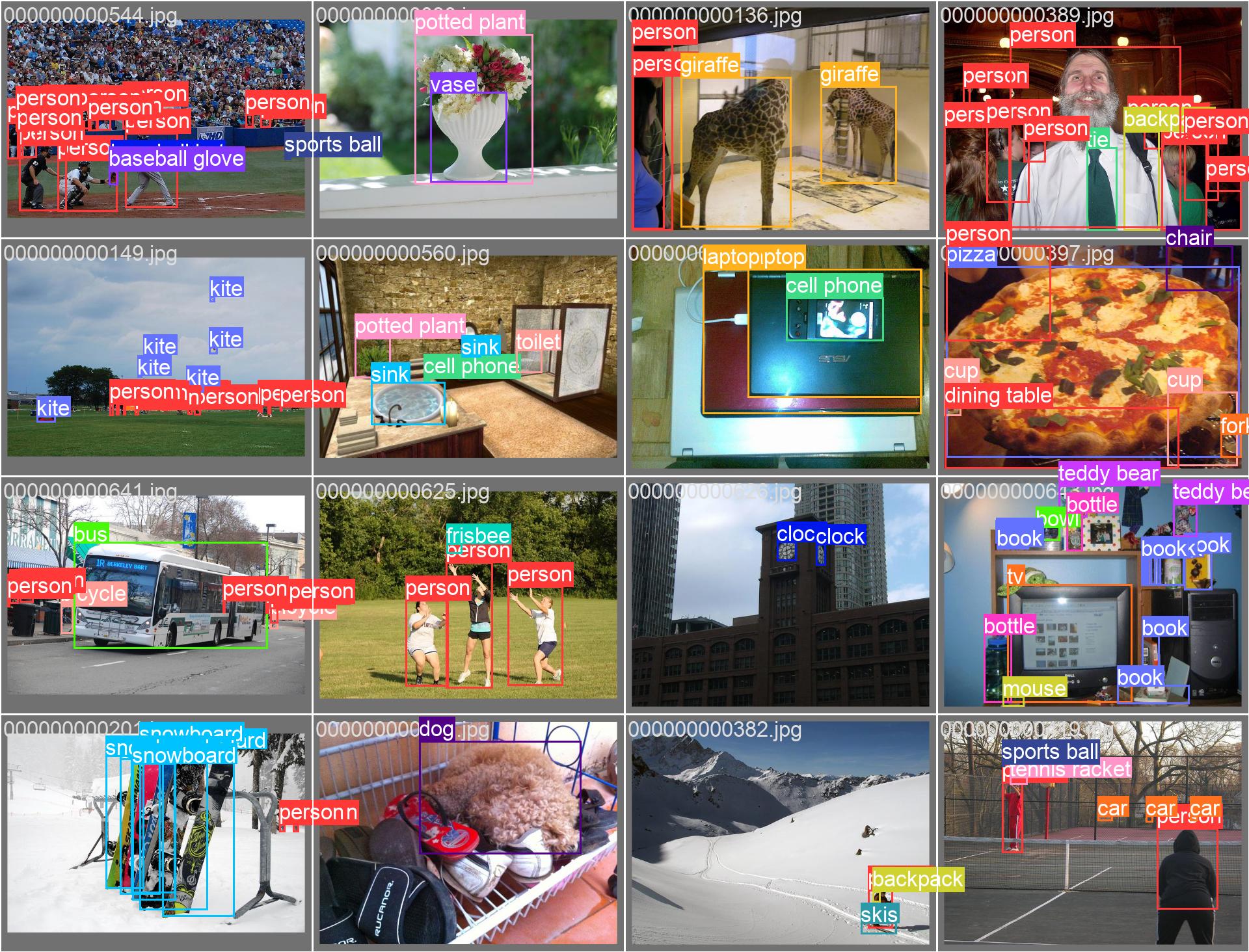}
      \caption{\textbf{Inference results for object detection.}}
  \label{fig:11}
\end{figure}

The inference results of different tasks are shown in Figure~\ref{fig:9}, \ref{fig:10} and \ref{fig:11}.

\end{document}